\documentclass[letterpaper]{article} 
\usepackage[draft]{aaai25}  
\usepackage{times}  
\usepackage{helvet}  
\usepackage{courier}  
\usepackage[hyphens]{url}  
\usepackage{graphicx} 
\usepackage{natbib}  
\usepackage{caption} 
\usepackage{algorithm}

\usepackage{booktabs}
\usepackage{color}
\usepackage{multirow}
\usepackage{amsmath}
\usepackage{algpseudocode}
\usepackage{longtable}

\algrenewcommand{\algorithmicrequire}{\textbf{Input:}}
\algrenewcommand{\algorithmicensure}{\textbf{Output:}}

\usepackage{newfloat}
\usepackage{listings}
\DeclareCaptionStyle{ruled}{labelfont=normalfont,labelsep=colon,strut=off} 
\floatstyle{ruled}
\newfloat{listing}{tb}{lst}{}
\floatname{listing}{Listing}
\title{LevelRAG: Enhancing Retrieval-Augmented Generation with\\Multi-hop Logic Planning over Rewriting Augmented Searchers}
\author {
    Zhuocheng Zhang\textsuperscript{\rm 1, 3},
    Yang Feng\textsuperscript{\rm 1, 2, 3}\thanks{ Corresponding author: Yang Feng.},
    Min Zhang\textsuperscript{\rm 4}
}
\affiliations{
    \textsuperscript{\rm 1}Key Laboratory of Intelligent Information Processing,\\
    Institute of Computing Technology, Chinese Academy of Sciences (ICT/CAS)\\
    \textsuperscript{\rm 2}Key Laboratory of AI Safety, Chinese Academy of Sciences\\
    \textsuperscript{\rm 3}University of Chinese Academy of Sciences, Beijing, China\\
    \textsuperscript{\rm 4}Harbin Institute of Technology (Shenzhen), Shenzhen, China\\
    zhangzhuocheng20z@ict.ac.cn, fengyang@ict.ac.cn, zhangminmt@hotmail.com
}

\usepackage{bibentry}

\begin{document}

\maketitle

\begin{abstract}

Retrieval-Augmented Generation (RAG) is a crucial method for mitigating hallucinations in Large Language Models (LLMs) and integrating external knowledge into their responses.
Existing RAG methods typically employ query rewriting to clarify the user intent and manage multi-hop logic, while using hybrid retrieval to expand search scope.
However, the tight coupling of query rewriting to the dense retriever limits its compatibility with hybrid retrieval, impeding further RAG performance improvements.
To address this challenge, we introduce a high-level searcher that decomposes complex queries into atomic queries, independent of any retriever-specific optimizations.
Additionally, to harness the strengths of sparse retrievers for precise keyword retrieval, we have developed a new sparse searcher that employs Lucene syntax to enhance retrieval accuracy.
Alongside web and dense searchers, these components seamlessly collaborate within our proposed method, \textbf{LevelRAG}. In LevelRAG, the high-level searcher orchestrates the retrieval logic, while the low-level searchers (sparse, web, and dense) refine the queries for optimal retrieval.
This approach enhances both the completeness and accuracy of the retrieval process, overcoming challenges associated with current query rewriting techniques in hybrid retrieval scenarios.
Empirical experiments conducted on five datasets, encompassing both single-hop and multi-hop question answering tasks, demonstrate the superior performance of LevelRAG compared to existing RAG methods.
Notably, LevelRAG outperforms the state-of-the-art proprietary model, GPT4o, underscoring its effectiveness and potential impact on the RAG field.


\end{abstract}

\begin{links}
    \link{Code}{https://github.com/ictnlp/LevelRAG}
\end{links}

\section{Introduction}

\begin{figure*}[t]
\centering
\includegraphics[width=\linewidth]{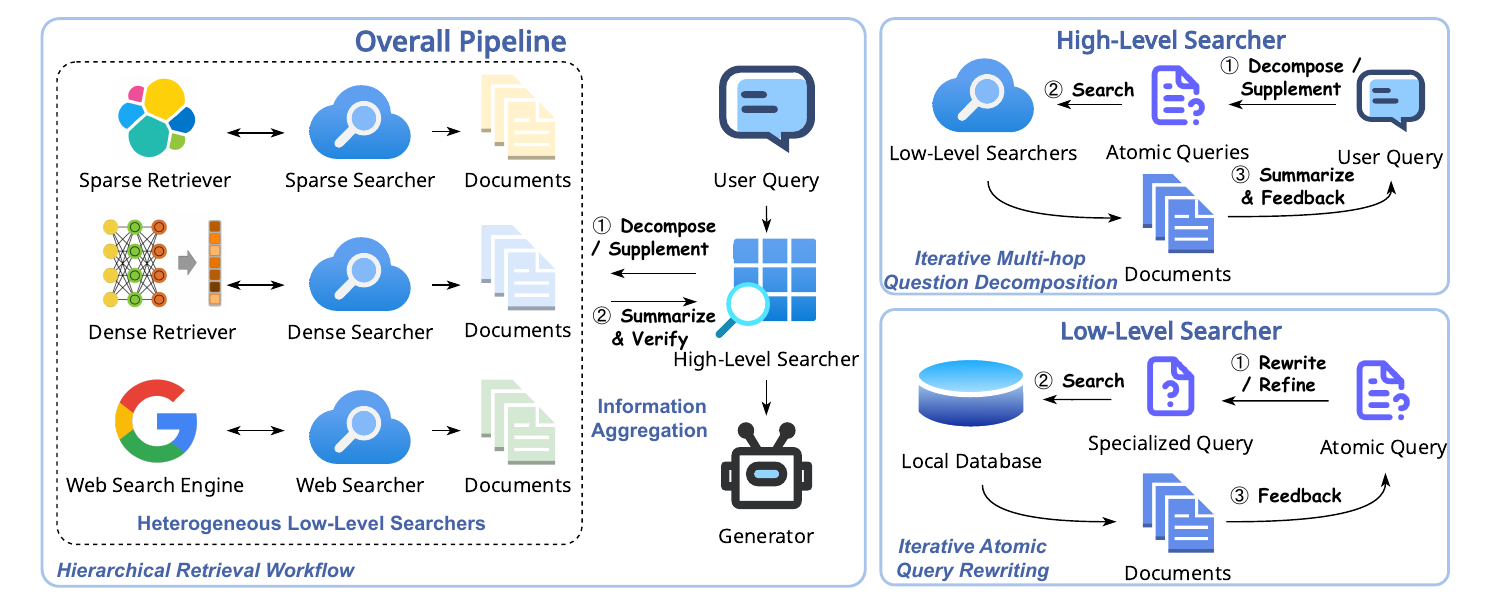}
\caption{Overview of the LevelRAG. The user query is initially processed by the high-level searcher. The decomposed atomic queries are handled by low-level searchers, which may rewrite and refine the queries before sending them into their corresponding retrievers. The retrieved documents are aggregated and summarized by the high-level searcher, then fed to the generator to generate the response. Both the high-level searcher and low-level searchers employs the feedback from the retrieved documents to refine or supplement their outputs.}
\label{fig:overview}
\end{figure*}

The introduction of scaling laws \cite{scaling_law,chinchilla} has led to significant advancements in Large Language Models (LLMs) \cite{gpt3,llama,llama2,llama3,phi3}, enabling them to achieve remarkable successes across a wide range of tasks \cite{bayling,llm4code,llm4opt,react,llm_subsume_retrieval}.
Despite their extensive training and vast knowledge base, LLMs continue to encounter challenges in accurately handling certain types of information, including less commonly known facts, specialized domain knowledge, and events that occurred after their pre-training phase \cite{when_not_trust_llm,hallucination_survey}.
In these cases, LLMs often fail to provide accurate responses and may even produce hallucinated answers.
Hence Retrieval-Augmented Generation (RAG) is proposed to enhance LLM-generated responses by appending real-time retrieval of relevant external knowledge to user input of LLMs, offering a promising solution to address these challenges \cite{rag_survey_1,rag_survey_2}.
However, due to the constraints of the search techniques and the coverage of databases, the retrieval results are often inaccurate and incomplete, affecting the effectiveness of RAG systems.
\par
Many researchers have made efforts to optimize RAG systems by improving the accuracy and completeness of retrieval results.
To improve accuracy, researchers have developed numerous query rewriting techniques \cite{query_rewrite,self_rag, rq_rag} which try to refine the query fed to a retriever to make the query better match with the retriever and better align to user intents.
On the other hand, to improve completeness, hybrid retrieval is employed to broaden the scope of retrieval, which integrates multiple retrievers and databases to maximize the advantages of diverse retrievers \cite{reacc,rapgen,bash_explainer,hybrid_code,hybrid_dense}.
\par
To enhance both the completeness and accuracy of the retrieval process, it is a good choice to combine the superiority of the two methods mentioned above. However, a key challenge arises from the fact that current query rewriting techniques are closely tied to the dense retriever, rendering them unsuitable for hybrid retrieval scenarios.
Meanwhile, to achieve complete retrieval results, we need to sort out all the retrieval results for hybrid retrieval to avoid overlapping and contradictory knowledge input to LLMs.
Besides, multi-hop planning for retrieval is also necessary as not all the user intents can be satisfied with one retrieval when fed with complex user input.
To address these limitations, we propose \textbf{LevelRAG}, a novel approach that decouples retrieval logic from retriever-specific rewriting to allow for greater flexibility for either function to achieve better expertise, and meanwhile involves interactions between them for global optimization.
As shown in Figure \ref{fig:overview}, LevelRAG is composed of a high-level searcher which is responsible for multi-hop logic planning along with information combing, and multiple low-level searchers which is the operator supporting query rewriting.
The high-level searcher conducts planning to decompose the user query into atomic queries for low-level searchers to retrieve, then summarizes the retrieved results of low-level searchers for information combing.
In the process of summarization, it also determines whether to invoke next-hop planning to supplement new atomic queries according to the completeness of summarized information for the user input.
The low-level searchers operate the real retrieval over databases with atomic queries from the high-level searcher, which may be rewritten to better fit the corresponding retriever.
Specifically, we employ a sparse searcher, a dense searcher, and a web searcher as the low-level searchers, among which the sparse searcher is good at exact retrieval for entities with Lucene syntax, the dense searcher is skilled at retrieval with complex queries which are transformed into pseudo-documents, and the web searcher is employed to leverage the vast knowledge from the internet to supplement the local database.
In conjunction with the high-level searcher, these searchers collectively guarantee the completeness and accuracy of the retrieval process.
\par
We conducted empirical experiments on five datasets, including both single-hop question answering (QA) tasks and multi-hop QA tasks. The experimental results demonstrate that LevelRAG significantly outperforms current RAG methods across these tasks. Notably, the response quality of our proposed method surpasses the state-of-the-art proprietary model, GPT4o, providing strong evidence for the superiority of our approach. Additionally, the mere utilization of our proposed sparse searcher outperforms numerous existing methods, which further reveals the effectiveness of our novel rewriting approach.
\par
In conclusion, our key contributions are as follows:
\begin{itemize}
    \item To address the challenge associated with the tight coupling of query rewriting to dense retrievers in hybrid retrieval scenarios, we propose LevelRAG, a method that integrates the high-level searcher with low-level searchers (sparse, web, and dense). This integration allows for optimal retrieval by refining queries across different retrievers, enhancing both the breadth and accuracy of the retrieval process.
    \item Targeting leveraging the strengths of sparse retrieval methods, particularly in precise keyword-based searches, we develop a new sparse searcher that leverages Lucene syntax to enhance retrieval accuracy.
\end{itemize}

\section{Related Works} 

\subsection{Query Rewriting}
Traditional RAG directly employs the user input as the query to the retriever to retrieve relevant information. However, the ambiguity of the user query and the misalignment between the query and the retriever prevent the retriever from retrieving supporting documents. To improve the retrieval accuracy, query rewriting techniques rephrase or extend the user query into a more precise and searchable query.
Specifically, \citet{query_rewrite} and \citet{rafe} proposed to enhance the query by training a specialized rewriting model.
\citet{hyde,query2doc} suggested that using pseudo contexts generated by large language models (LLMs) as queries can significantly improve the relevance of retrieved documents. Compared to the original user query, pseudo contexts contain more relevant information, thereby increasing the semantic overlap between the supporting documents and the initial query. To further refine the query, iterative refinement was introduced by \citet{itrg,ircot,iter_retgen,blendfilter} to employ the retrieved contexts as feedback. Additionally, query decomposition was proposed by \citet{rq_rag} to convert the complex multi-hop question into multiple searchable sub-questions. 

\subsection{Hybrid Retrieval}
Different types of retrievers excel at retrieving specific kinds of information. For instance, sparse retrievers are particularly effective at identifying relevant content based on explicit entities mentioned in the query. Dense retrievers, on the other hand, are adept at handling fuzzy queries, and capturing semantic meanings. Web search engines leverage the vast breadth of information available online, offering extensive knowledge retrieval. Consequently, employing a combination of these retrievers allows for a more comprehensive search strategy, capitalizing on the strengths of each method and broadening the scope of retrieval.
Specifically, \citet{reacc,rapgen,bash_explainer,hybrid_code} employed both BM25 retriever and dense retriever to retrieve the relevant snippets for better code generation. \citet{unims_rag} proposed leveraging large language models to route information from various sources, thereby enhancing the effectiveness of personalized knowledge-based dialogue systems. Different from simply concatenating all the retrievers, \citet{crag} employed a web search engine as the supplement of the dense retriever. Ensemble of Retrievers (EoR), introduced by \citet{hybrid_dense}, leverages ensemble learning to optimize collaboration among multiple retrievers. 

\subsection{Concurrent Work}
Similar to our proposed LevelRAG, MindSearch \cite{mind_search} also employs a two-level retrieval architecture. At the upper level, MindSearch utilizes WebPlanner, which decomposes the user query into a directed acyclic graph (DAG). The lower level, known as WebSearcher, performs hierarchical information retrieval using web search engines, gathering relevant information for the WebPlanner. While MindSearch shares similarities with our proposed LevelRAG, three key distinctions set them apart:

\begin{itemize}
    \item The motivation is different. MindSearch is designed to address the challenge of adapting the vast amount of web information to the limited context window of LLMs during internet searches. In contrast, LevelRAG is motivated by the need to overcome limitations in current query rewriting techniques that are inadequate for hybrid retrieval systems.
    \item The methods differ significantly in their implementation. MindSearch employs a code interpreter to decompose user queries, whereas LevelRAG leverages natural language processing for the decomposition process. Additionally, while MindSearch incorporates only a WebSearcher at the lower level, LevelRAG is capable of integrating multiple low-level searchers to gather valuable information.
    \item The contributions are fundamentally different. MindSearch introduces a multi-agent framework for complex web information-seeking and integration tasks, whereas our work presents a novel approach that combines the strengths of both hybrid retrieval and query rewriting techniques. Additionally, to the best of our knowledge, our proposed sparse searcher is the first rewriting framework designed for sparse retrievers.
\end{itemize}

\section{LevelRAG}
LevelRAG achieves accurate retrieval through close collaboration between a high-level searcher and several low-level searchers. The high-level searcher is responsible for planning which sub-questions need to be retrieved to meet the user intent, while the low-level searchers, in collaboration with its retriever, gather relevant information for these sub-questions as comprehensively as possible. In this section, we first provide a detailed explanation of the high-level searcher, followed by a description of each low-level searcher individually.

\begin{figure}[t]
\centering
\includegraphics[width=\columnwidth]{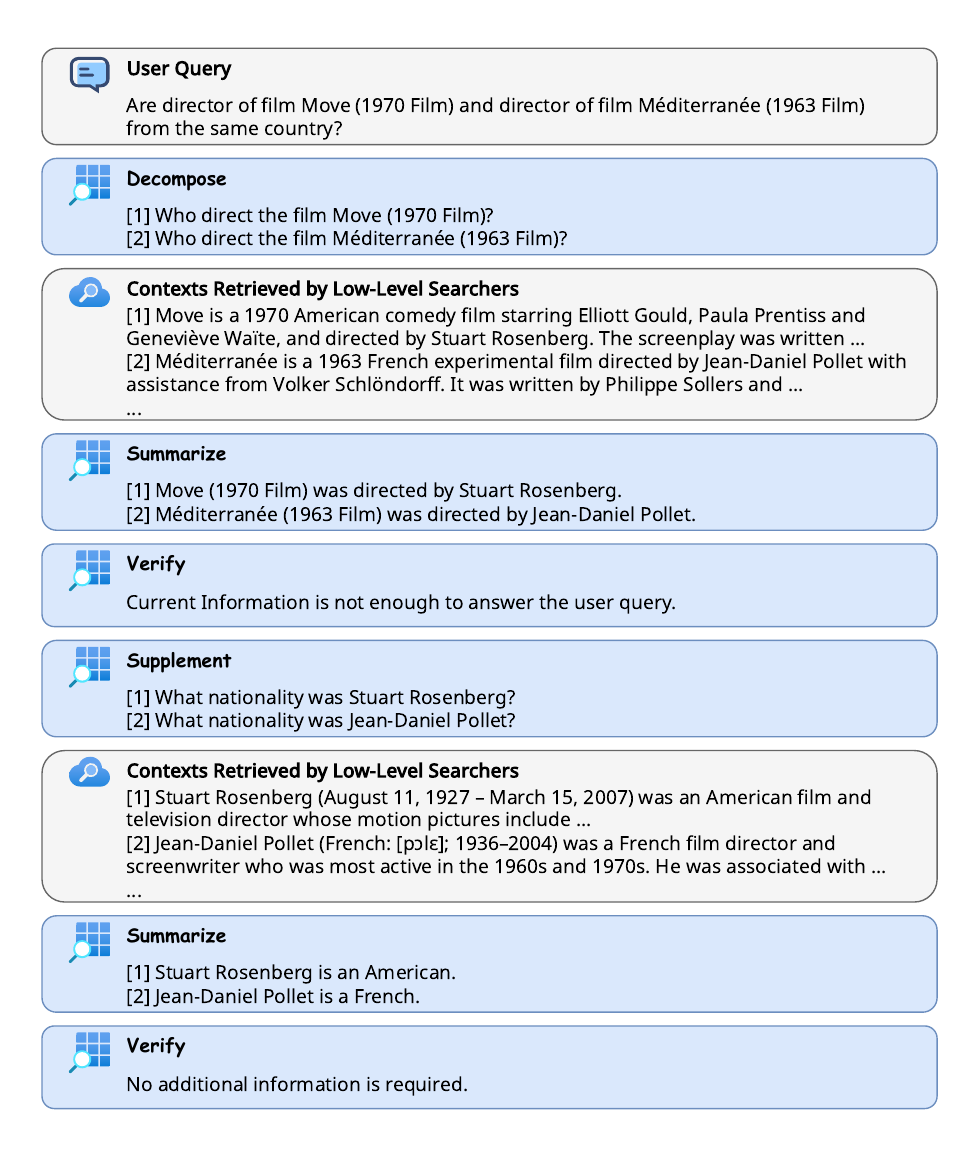}
\caption{An example of how the high-level searcher processes a user query. The high-level searcher performs four key actions: \textit{decompose}, \textit{summarize}, \textit{verify}, and \textit{supplement}. Actions within blue boxes are carried out by the high-level searcher, while those within grey boxes are performed by the low-level searchers and user.}
\label{fig:highlevel_example}
\end{figure}

\subsection{\label{sec:high_level}High-Level Searcher}
The high-level searcher orchestrates the overall search process, managing low-level searchers by decomposing the original query into atomic queries.
Before delving into the retrieval process, we will first introduce four key actions taken by the high-level searcher:
\begin{itemize}
    \item \textit{Decompose}: The high-level searcher breaks down the user query $q$ into simpler, more specific atomic queries $\mathbf{\overline{q}}=[\overline{q}_1,...,\overline{q}_m]$ that are easier to address individually. $m$ represent the number of atomic queries, which is determined automatically by the high-level searcher.
    \item \textit{Summarize}: After retrieving relevant documents $\mathbf{D}_i=[D^1_i,...,D^k_i]$ for atomic query $\overline{q}_i$, the high-level searcher condenses them into a brief summary $s_i$ by directly answering the atomic query $\overline{q}_i$.
    \item \textit{Verify}: The high-level searcher checks whether the summarized information $\mathbf{s}=[s_1,...,s_m]$ is sufficient to answer the original user query. If the information is not enough, this step triggers further supplementation.
    \item \textit{Supplement}: As some specific information needs becomes apparent only after the preliminary atomic questions being answered, the high-level searcher identifies additional atomic queries to fully address the user query based on the summaries $\mathbf{s}$.
\end{itemize}

Figure \ref{fig:highlevel_example} illustrates the overall retrieval pipeline of the high-level searcher when dealing with a complex user query.
The high-level searcher begins by taking the complex user query $q$ as input and \textit{decomposes} it into multiple single-hop queries $\mathbf{\overline{q}}$.
Each low-level searcher then interacts with its corresponding retriever to collect the most relevant documents $\mathbf{D}_i$ for the atomic query $q_i$.
Due to the large volume of retrieved documents and the potential noise they contain, concatenating them directly into the prompt for subsequent processes may result in context length overflow and the "lost-in-the-middle" phenomenon.
To address this, the high-level searcher performs a \textit{summarize} operation of these documents.
However, we observed that directly summarizing these contexts can lead to the loss of information crucial for answering the user query.
Therefore, we implemented a strategy that summarizes the context by answering the atomic query.
This approach ensures the conciseness of the contexts while preserving the information necessary for accurately addressing the user query.
Subsequently, the summarized contexts are then aggregated to \textit{verify} their sufficiency in addressing the original user query. If the collated summaries provide comprehensive coverage, they are used as the context for the generator to generate the final response.
Otherwise, the high-level searcher will \textit{supplement} new atomic queries to retrieve further information.
The process iterates until the summaries are comprehensive enough to answer the user query.

\subsection{\label{sec:low_level}Low-Level Searchers}
Low-level searchers are employed to retrieve relevant documents for each atomic query. Since low-level searchers interact directly with their corresponding retrievers, the accuracy of retrieval can be further improved by aligning the query with the specific retriever.
In this paper, we implement three low-level searchers: the sparse searcher, the dense searcher, and the web searcher.

\begin{algorithm}[tb]
\caption{Sparse Search Pipeline}
\label{alg:sparse_searcher}
\begin{algorithmic}[1] 
\Require Atomic query $\overline{q}$
\Require Maximum rewrite times $N$
\Ensure Retrieved documents \mbox{$\mathbf{D}=[d_1,\ldots,d_k]$}
\State \textbf{Initialize} $\mathbf{\mathcal{Q}}=\emptyset$.
\State \textbf{Initialize} $n=0$
\State $q_0=\textbf{Rewrite}(\overline{q})$.
\State \textbf{Add} $\tilde{q}_0$ to $\mathcal{Q}$.
\While{$n<N$} 
  \State Let $\tilde{q}=\text{POP}(\mathcal{Q})$.
  \State Let $\mathbf{D}=\text{Retrieve}(\tilde{q})$.
  \If {\textbf{Verify}($\mathbf{D}$, $\overline{q}$)}
    \State \textbf{return} $\mathbf{D}$.
  \EndIf
  \State $\tilde{q}_1=\textbf{Extend}(\tilde{q}, \mathbf{D})$.
  \State $\tilde{q}_2=\textbf{Emphasize}(\tilde{q}, \mathbf{D})$.
  \State $\tilde{q}_3=\textbf{Filter}(\tilde{q}, \mathbf{D})$.
  \State \textbf{Add} $\tilde{q}_1$ to $\mathcal{Q}$.
  \State \textbf{Add} $\tilde{q}_2$ to $\mathcal{Q}$.
  \State \textbf{Add} $\tilde{q}_3$ to $\mathcal{Q}$.
\EndWhile
\end{algorithmic}
\end{algorithm}

\subsubsection{Sparse Searcher.}
Sparse retrievers offer the distinct advantage of precisely retrieving keywords such as named entities. To further leverage this capability, we leverage the Lucene syntax, a powerful and widely supported query language for searching and indexing text data.
Lucene offers a rich set of operators and features that enable precise and flexible searches across large text datasets.
Based on this, we introduce a novel iterative query rewriting process for the sparse retriever.
As shown in Algorithm \ref{alg:sparse_searcher}, the atomic query is first reformulated into retriever-friendly keywords and enqueued in the query queue $\mathcal{Q}$. A breadth-first search (BFS) is then initiated to further refine this query.
During this process, queries are dequeued from the queue $\mathcal{Q}$ at a time, and the sparse retriever is employed to retrieve relevant documents $\mathbf{D}$.
Similar to the high-level searcher, the sparse searcher will assess the sufficiency of the retrieved documents.
Once the retrieved documents are sufficient to address the atomic query $\overline{q}$, they are returned to the high-level searcher. Otherwise, three feedback operations will be applied to the current query $\tilde{q}$, including \textit{extend}, \textit{emphasize}, and \textit{filter}.
\begin{itemize}
    \item \textit{extend}: The sparse searcher appends an additional keyword to the query $\tilde{q}$, thereby broadening the scope of retrieval. To protect the keyword from being incorrectly processed by the tokenizer inside the sparse retriever, we employ the quotation operator to mark the new keyword.
    \item \textit{filter}: To filter out noisy documents, the sparse searcher appends a filter word to the query using the Lucene operator "\texttt{-}".
    \item \textit{emphasize}: This operation first selects a specific keyword in the current query $\tilde{q}$. Then the Lucene operator "\texttt{\^{}}" is adopted to increase the weight of this keyword.
\end{itemize}
These enhanced queries will be added to the query queue $\mathcal{Q}$ for further retrieval. This process continues until the retrieved documents are comprehensive enough or the maximum number of loop $N$ is reached.

\subsubsection{Dense Searcher.}
Compared to sparse retrievers, dense retrievers are more adept at capturing semantic information within queries. To further leverage this strength, we implemented a query rewriting process inspired by ITRG \cite{itrg} and HyDE \cite{hyde}, which iteratively enriches the semantic content of the query.
Specifically, our approach involves constructing a pseudo-document to augment the semantic details of the query, particularly when the original query fails to retrieve relevant documents.
If the context retrieved using the enhanced query still fails to adequately address the atomic query, a new pseudo-document will be constructed based on the retrieved information.
Similar to the sparse searcher, this iterative process continues until the retrieved information is sufficient to answer the atomic question.

\subsubsection{Web Searcher.}
With the development of the internet, search engines have become indispensable tools for people. To enhance the quality of search results, major companies have undertaken extensive optimizations of search engines. Given that search engines inherently offer strong performance and that their APIs are relatively expensive, in our Web Searcher, we directly input atomic queries into the search engine and return the abstracts provided by the search engine as the context for the high-level searcher.

\begin{table*}[t]
\resizebox{\linewidth}{!}{%
\begin{tabular}{lccccccccccccccc}
\toprule
\multicolumn{1}{c}{\multirow{2}{*}{Methods}}        & \multicolumn{3}{c}{PopQA}  & \multicolumn{3}{c}{NQ}                     & \multicolumn{3}{c}{TriviaQA}               & \multicolumn{3}{c}{HotpotQA}               & \multicolumn{3}{c}{2WikimultihopQA}        \\
\multicolumn{1}{c}{}                & Succ         & Acc          & F1           & Succ         &     EM       & F1           & Succ         &    Acc       &    F1        & Succ         & Acc          &  F1          & Succ         & Acc          & F1           \\
\cmidrule(lr){1-16}
\multicolumn{16}{l}{\textit{Generation without Retrieval}}\\
\cmidrule(lr){1-16}
Qwen2 7B     \cite{qwen2}           &       -      &    25.80     &     21.53    &       -      &     15.21    &     23.42    &       -      &    46.65     &     47.66    &      -       &     21.80    &     27.76    &       -      &     33.12    &     32.58    \\
Qwen2 72B    \cite{qwen2}           &       -      &    24.18     &     23.42    &       -      &     30.00    &     41.34    &       -      &    77.26     &     73.74    &      -       &     31.94    &     39.38    &       -      &     37.52    &     38.08    \\
Llama3.1 8B  \cite{llama3}          &       -      &    25.38     &     21.96    &       -      &     22.74    &     34.25    &       -      &    63.82     &     64.50    &      -       &     35.58    &     31.61    &       -      &     24.27    &     29.10    \\
Llama3.1 70B \cite{llama3}          &       -      &    31.45     &     28.70    &       -      &     28.09    &     44.31    &       -      &    77.26     &     78.30    &      -       &     33.38    &     39.18    &       -      &     39.27    &     35.27    \\
GPT4o-mini   \cite{instruct_gpt}    &       -      &    31.81     &     26.47    &       -      &     27.20    &     41.10    &       -      &    73.31     &     73.49    &      -       &     33.40    &     38.62    &       -      &     35.15    &     33.76    \\
GPT4o        \cite{instruct_gpt}    &       -      &    45.96     &     40.61    &       -      &     28.61    &     44.37    &       -      &    82.53     &\textbf{82.66}&      -       &\textbf{43.93}&     48.58    &       -      &     49.64    &     44.50    \\
\cmidrule(lr){1-16}
\multicolumn{16}{l}{\textit{Generation with Retrieval}}\\
\cmidrule(lr){1-16}
Self-RAG 7B  \cite{self_rag}        &       -      &     54.90    &     -        &       -      &       -      &     -        &       -      &    66.40     &       -      &      -       &       -      &       -      &       -      &      -       &      -       \\
Self-RAG 13B \cite{self_rag}        &       -      &     55.80    &     -        &       -      &       -      &     -        &       -      &    69.30     &       -      &      -       &       -      &       -      &       -      &      -       &      -       \\
ITRG 7B*     \cite{itrg}            &     43.86    &     41.10    &     34.65    &     55.03    &     20.64    &     33.32    &     71.69    &    67.21     &     60.80    &     43.14    &     35.69    &     37.14    &     38.23    &     33.37      &     33.45    \\
IR-COT 7B*   \cite{ircot}           &     41.12    &     39.81    &     26.96    &     47.84    &     16.43    &     25.92    &     65.90    &    58.91     &     39.33    &     41.17    &     33.80    &     22.75    &     40.95    &     35.81      &     19.72    \\
RQ-RAG 7B    \cite{rq_rag}          &       -      &     -        &     57.10    &       -      &       -      &     -        &       -      &     -        &       -      &      -       &       -      &\textbf{62.60}&       -      &      -       &     44.80    \\
ReSP 8B      \cite{resp}            &       -      &     -        &     -        &       -      &       -      &     -        &       -      &     -        &       -      &      -       &       -      &     38.30    &       -      &      -       &     47.20    \\
ChatQA 8B    \cite{chatqa}          &       -      &     59.80    &     -        &       -      &     42.40    &     -        &       -      &     87.60    &       -      &      -       &       -      &     44.60    &       -      &      -       &     31.90    \\
ChatQA 70B   \cite{chatqa}          &       -      &     58.30    &     -        &       -      &     47.00    &     -        &       -      &     91.40    &       -      &      -       &       -      &     54.40    &       -      &      -       &     37.40    \\
RankRAG 8B   \cite{rank_rag}        &       -      &     64.10    &     -        &       -      &     50.60    &     -        &       -      &     89.50    &       -      &      -       &       -      &     46.70    &       -      &      -       &     36.90    \\
RankRAG 70B  \cite{rank_rag}        &       -      &     65.40    &     -        &       -      &\textbf{54.20}&     -        &       -      &\textbf{92.30}&       -      &      -       &       -      &     55.40    &       -      &      -       &     43.90    \\
Qwen2 7B + BM25                     &     59.04    &     51.32    &     47.57    &     58.37    &     30.72    &     39.36    &     75.40    &     64.80    &     66.72    &     52.19    &     35.41    &     43.53    &     57.72    &     43.66    &     44.79    \\
Qwen2 7B + Dense                    &     83.06    &     65.83    &     63.93    &     73.74    &     37.01    &     47.30    &     78.12    &     66.58    &     68.81    &     49.32    &     32.88    &     40.65    &     49.03    &     39.39    &     40.47    \\
Qwen2 7B + Web                      &     60.97    &     50.82    &     48.97    &     68.45    &     36.73    &     48.65    &     76.86    &     70.80    &     73.17    &     38.81    &     29.70    &     37.54    &     38.20    &     35.35    &     36.21    \\
\cmidrule(lr){1-16}
\multicolumn{16}{l}{\textit{Our Proposed methods}}\\
\cmidrule(lr){1-16}
Sparse Searcher 7B                  &      73.84   &     63.26    &     61.77    &    61.41     &     32.55    &     41.89    &     74.92    &     65.44    &     67.35    &     52.36    &     36.07    &     44.38    &     58.34    &     48.31    &    48.84     \\
Dense Searcher 7B                   &      77.06   &     62.19    &     60.23    &    74.93     &     38.81    &     49.35    &     79.35    &     69.06    &     71.40    &\textbf{53.46}&     35.69    &     43.86    &     53.81    &     44.92    &    45.59     \\
LevelRAG 7B                         &\textbf{85.42}&\textbf{66.98}&\textbf{65.52}&\textbf{86.51}&     41.58    &\textbf{53.24}&\textbf{88.23}&     76.12    &     78.21    &     51.22    &     43.78    &     52.28    &\textbf{73.18}&\textbf{70.50}&\textbf{69.33}\\
\bottomrule
\end{tabular}%
}
\caption{Results of our proposed LevelRAG and baselines on single-hop and multi-hop tasks. Results unavailable in public reports are marked as "-". The "*" mark indicates the results are reproduced using FlexRAG\footnotemark[9] with Qwen2 7B as the generator. \textbf{Bold} number indicates the best performance among all methods. Succ, Acc, F1 and EM represent the success rate of retrieval, the accuracy of responses, the F1 score of responses, and the Exact Match metrics, respectively.}
\label{tab:main}
\end{table*}

\section{Experimental Settings}
\subsection{Evaluation Tasks and metrics}
We conduct extensive experiments across 5 widely used knowledge-intensive QA tasks, including three single-hop QA tasks and two multi-hop QA tasks. For the single-hop QA tasks, we utilize PopQA \cite{popqa}, Natual Questions (NQ) \cite{nq}, and TriviaQA \cite{triviaqa}. Following previous work \cite{self_rag}, we use the long-tail subset of the PopQA and the full test set of NQ and TriviaQA. For multi-hop QA tasks, we employ HotpotQA \cite{hotpot} and 2WikimultihopQA \cite{2wiki} development sets. All these datasets are collected from the FlashRAG repository\footnote{https://huggingface.co/datasets/RUC-NLPIR/FlashRAG\_datasets}. Our experiments are conducted
\par
To evaluate performance across these tasks, we use retrieval success rate, response accuracy, and response F1 score to assess both the retrieval process and response quality, with the exception of the NQ dataset. Both retrieval success rate and response accuracy are computed based on whether gold answers are included in the text. For NQ, in line with established practices \cite{flash_rag,rank_rag}, we substitute response accuracy with Exact Match (EM), as EM is widely used in prior studies. Following \citet{flash_rag}, we lowercase all the responses before calculating the metrics.

\subsection{Baselines}
We evaluate our approach against a wide range of strong baselines, which we categorize into two main groups: generation without retrieval and generation with retrieval. For the generation without retrieval group, we consider both leading open-source model series, Llama 3.1 \cite{llama3} series and Qwen 2 \cite{qwen2} series, as well as proprietary models from OpenAI. In the generation with retrieval group, we compare our method with several advanced approaches, including ITRG \cite{itrg} and IR-COT \cite{ircot}, which employ iterative refinement to optimize query performance; SelfRAG \cite{self_rag}, which performs adaptive retrieval based on the reflection token; RQ-RAG \cite{rq_rag}, which utilizes query decomposition to enhance performance in multi-hop QA tasks; ReSP \cite{resp}, which incorporates a summarizer to improve iterative retrieval process; and two specialized RAG models, namely ChatQA and RankRAG \cite{chatqa,rank_rag}.
\par
Additionally, we conduct experiments using the vanilla RAG method, where contexts are retrieved directly through a sparse retriever, a dense retriever, and a web search engine, respectively. As \citet{llm_subsume_retrieval,long_rag} demonstrated that current LLMs have a strong ability to select the supporting facts from the noisy contexts, we concatenate all the retrieved documents into the prompt to generate the final response.

\subsection{Implementation Details of LevelRAG}
In our primary experiments, we employ Qwen2 7B as the base model of our LevelRAG. For models with a parameter fewer than 10 billion, we deploy them using VLLM\footnote{https://github.com/vllm-project/vllm} on two NVIDIA RTX 3090 GPUs. For models with more than 70 billion parameters, we utilize Ollama\footnote{https://github.com/ollama/ollama} to deploy the quantized models on two NVIDIA L40 GPUs. In all our experiments, we set the temperature to zero to ensure the reproducibility of the results.
\par
For the sparse retriever, we build the index using ElasticSearch\footnote{https://github.com/elastic/elasticsearch}. For the dense retriever, we employ the contriever\footnote{https://huggingface.co/facebook/contriever-msmarco} \cite{contriever} finetuned on MS-MARCO to encode the corpus and SCaNN\footnote{https://github.com/google-research/google-research/tree/master/scann} to index the encoded embeddings. Following the previous study \cite{self_rag}, we use the pre-processed Wikipedia corpus\footnote{https://github.com/facebookresearch/atlas} processed by \citet{atlas} where all the Wikipedia pages are split into chunks of 100 words.
We utilize the Bing Web Search API provided by Azure\footnote{https://azure.microsoft.com/} as our web search engine. Unless otherwise specified, we retrieve 10 documents per query for all retrievers. In the case of the sparse searcher, we limit the breadth-first search (BFS) depth to 3, allowing for a maximum of 27 refinements. For the dense searcher, we restrict the number of query rewrites to a maximum of 3.

\section{Results and Analysis}

\footnotetext[9]{https://github.com/ictnlp/flexrag}

\subsection{Main Results}
As shown in Table \ref{tab:main}, our proposed LevelRAG achieves the best overall performance when compared to the methods with similar model sizes.
Specifically, LevelRAG surpasses the state-of-the-art method by 1.58 and 22.13 on PopQA and 2WikimultihopQA respectively when using the F1 metric.
\par
As for non-retrieval models, we notice that the performance of the model improves significantly as the scale of the parameters increases. Specifically, GPT4o surpasses several retrieval-based methods.
When compared with non-retrieval models, including GPT4o, LevelRAG demonstrates a substantial performance advantage, with the exception of the TriviaQA dataset.
However, we found that Llama3.1 70B and GPT4o can correctly answer more than 75\% of the questions in the TriviaQA dataset, which is even higher than the retrieval success rate of the vanilla RAG system. This result clearly indicates that generating the response for this dataset does not heavily rely on external knowledge.
In addition, the relatively lower performance of the Qwen2 7B, the base model of LevelRAG, is also an important reason why our method fails to outperform all the non-retrieval models on this dataset.
\par
When compared with retrieval-based methods, the F1 score of LevelRAG surpasses all other methods across the PopQA, NQ, and 2WikimultihopQA datasets. Notably, when compared to RankRAG 70B, the strongest specialized model we could find, LevelRAG achieves comparable performance while utilizing only one-tenth of the parameters.
Moreover, the response accuracy and the F1 score of LevelRAG on the 2WikimultihopQA significantly outperform those of the state-of-the-art method, ReSP, which also leverages the feedback to refine the query. These results underscore the effectiveness of our high-level searcher in addressing complex multi-hop reasoning tasks.
However, the failure of our approach to achieving the best results on TriviaQA also reveals a weakness of our approach: the lack of adaptive mechanisms to determine when retrieval is necessary.
\par
Our proposed method demonstrates a significant improvement in retrieval success rates compared to the vanilla RAG system, which directly uses the user query to retrieve context documents.
Specifically, for single-hop QA tasks, the sparse searcher outperforms the naive BM25 search by 14.8 points on PopQA and 1.83 points on NQ.
Similarly, the dense searcher enhances the retrieval success rate by 1.19 and 1.23 points on NQ and TriviaQA, respectively. 
Furthermore, when the retrieved documents are aggregated by the high-level searcher, the retrieval success rate shows additional improvement.
For the 2WikimultihopQA dataset, LevelRAG demonstrates a significantly higher retrieval success rate compared to using only the low-level searcher, underscoring the effectiveness of the high-level searcher. Notably, although the retrieval success rate of LevelRAG  is lower than that of both the sparse and dense searchers, its response accuracy and F1 score surpass those of the other methods. We attribute this discrepancy to the summarization process employed by the high-level searcher, which, while potentially leading to some information loss, also effectively reduces noise in the context, thereby enhancing the overall quality of the responses.

\subsection{Ablation Study}

\begin{table}[t]
\centering
\begin{tabular}{lccc}
\toprule
Operations             & Succ         & F1           & Acc          \\
\cmidrule(lr){1-1}  \cmidrule(lr){2-4}
vanilla BM25           & 44.50        & 40.26        & 38.10        \\
+ \textit{decompose}   & 56.90        & 41.44        & 40.10        \\
+ \textit{summarize}   & 41.00        & 42.08        & 39.90        \\
+ \textit{supplement}  & 62.80        & 52.83        & 51.90        \\
\bottomrule
\end{tabular}%
\caption{The ablation studies for key executed by the high-level searcher on the first 1,000 samples of 2WikimultihopQA dataset.}
\label{tab:ablation_high_level}
\end{table}

\begin{table}[t]
\centering
\begin{tabular}{lccc}
\toprule
Operations             & Succ         & F1           & Acc          \\
\cmidrule(lr){1-1}  \cmidrule(lr){2-4}
vanilla BM25           & 63.40        & 47.90        & 53.10        \\
+ \textit{rewrite}     & 73.70        & 58.21        & 60.30        \\
+ \textit{feedback}    & 75.30        & 59.50        & 61.70        \\
\bottomrule
\end{tabular}%
\caption{The ablation studies for key operations executed by the sparse searcher on the first 1,000 samples of PopQA dataset.}
\label{tab:ablation_low_level}
\end{table}

\subsubsection{High-level searcher.}
To further analyze the effectiveness of each operation in our proposed high-level searcher, we conduct a series of experiments on the first 1,000 samples of the 2Wikimultihopqa dataset. To exclude the influence of low-level searchers, we do not use the low-level searchers we proposed but instead directly utilize ElasticSearch to search for the atomic query generated by the high-level searcher.
As shown in Table \ref{tab:ablation_high_level}, we start by using BM25 directly and gradually add \textit{decompose}, \textit{summarize}, and \textit{supplement} operations to the searcher to observe the impact of these operations. The experimental results demonstrate that both the \textit{decompose} and \textit{supplement} operations increase the retrieval success rate significantly.
However, by applying the \textit{summarize} operation, the retrieval success rate decreases. This decline is partly because the \textit{summarize} operation still loses some information, and partly because there are more false positive examples of success rate in longer contexts.
Notably, the F1 score even improved after applying the \textit{summarize} operation, which clearly demonstrates that the \textit{summarize} operation is beneficial for subsequent processes.

\subsubsection{Sparse searcher.}
Similarly, we also conduct an experiment on the first 1,000 samples of the PopQA dataset to analyze the effectiveness of key operations executed by our sparse searcher. As shown in Table \ref{tab:ablation_low_level}, we mainly explore the impact of the two key operations, \textit{rewrite} and \textit{feedback}, where the former modifies the atomic queries into keywords, while the latter further refines the keywords based on the retrieval results. The experimental results show that both operations are beneficial in improving the retrieval success rate and response quality. Specifically, rewrite contributes more to improving the retrieval success rate.

\subsection{The Impact of The Base Model}

\begin{table}[t]
\centering
\begin{tabular}{lccc}
\toprule
Base Model             & Succ         & F1           & Acc          \\
\cmidrule(lr){1-1}  \cmidrule(lr){2-4}
Qwen2 7B               &  62.80       & 52.83        & 51.90        \\
Qwen2 72B              &  75.40       & 65.28        & 68.20        \\
GPT4o mini             &  69.60       & 60.28        & 63.20        \\
\bottomrule
\end{tabular}%
\caption{The impact of the base model in high-level searcher. The experiments are conducted on the first 1,000 samples of the 2WikimultihopQA dataset.}
\label{tab:basemodel_high_level}
\end{table}

\begin{table}[t]
\centering
\begin{tabular}{lccc}
\toprule
Base model             & Succ         & F1           & Acc          \\
\cmidrule(lr){1-1}  \cmidrule(lr){2-4}
Qwen2 7B               &  75.30       &  59.50       & 61.70        \\
Qwen2 72B              &  74.50       &  58.99       & 63.00        \\
GPT4o mini             &  77.00       &  60.68       & 65.40        \\
\bottomrule
\end{tabular}%
\caption{The impact of the base model in sparse searcher. The experiments are conducted on the first 1,000 samples of the PopQA dataset.}
\label{tab:basemodel_low_level}
\end{table}

To figure out the impact of the selection of the base model on LevelRAG, we conduct experiments on 2WikimultihopQA and PopQA tasks. Similar to the previous section, we use the first 1,000 samples from each dataset to reduce the inference cost. As shown in Table \ref{tab:basemodel_high_level}, switching the base model from Qwen2 7B to either Qwen2 72B or GPT4o mini can improve the performance of the high-level searcher significantly. Notably, using Qwen2 72B yields better results than using GPT-4o mini, indicating that the capability of the base model is an important factor affecting the performance of our high-level searcher.
\par
Similarly, employing Qwen2 72B and GPT4o mini can also improve the overall performance of the sparse searcher. However, in this case, the improvements from using GPT-4o mini and Qwen2 72B are both relatively small, suggesting that the sparse searcher we proposed is not sensitive to the base model capability. The reason for this difference is that the high-level searcher needs to control the search logic, making the reasoning ability of the base model more critical. In contrast, the tasks of the sparse searcher are relatively simple, and it can achieve good results without using a more powerful base model.

\section{Conclusion}
In this work, we propose LevelRAG, a hierarchical approach that contains a high-level searcher and several low-level searchers. In LevelRAG, the high-level searcher plans the multi-hop logic while low-level searchers optimize the query to align with their corresponding retrievers. Our experimental results demonstrate that LevelRAG not only surpasses all counterparts across three single-hop QA tasks but also exhibits superior performance in complex multi-hop QA scenarios. Further analysis shows the effectiveness of each component in our high-level searcher and sparse searcher. We hope this work can contribute to the development of retrieval-augmented generation.

\bigskip

\bibliography{aaai25}

\appendix
\newpage
\section{Prompts}
\begin{figure}[h]
    \centering
    \includegraphics[width=\columnwidth]{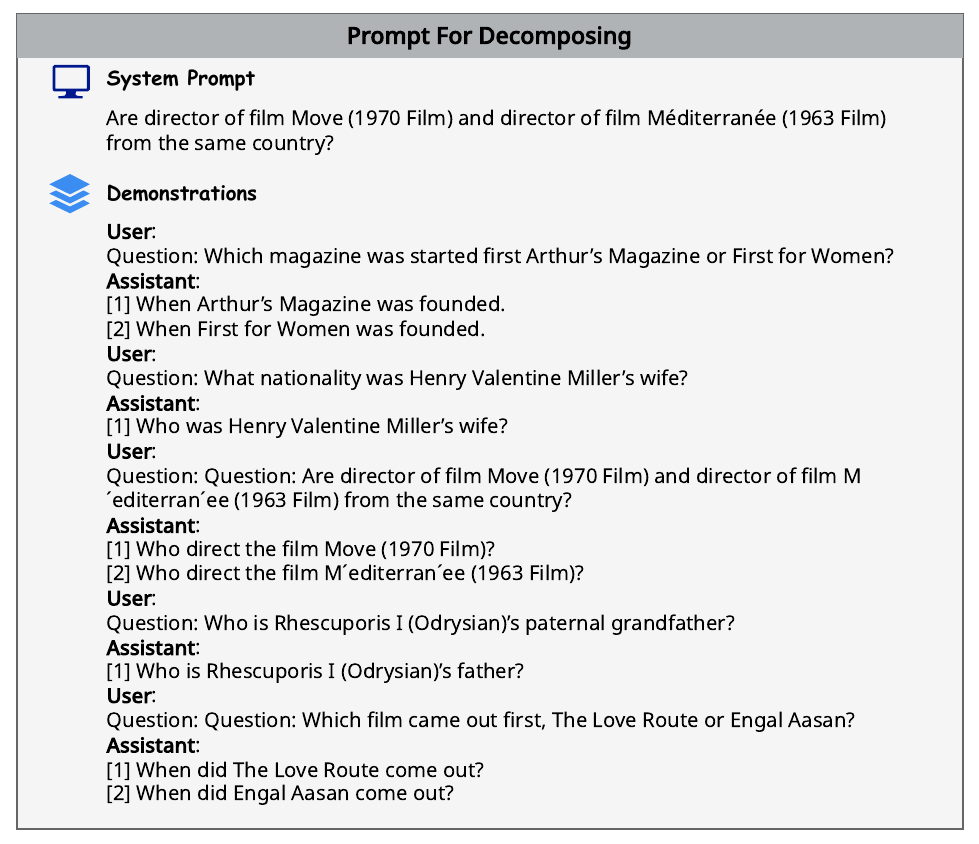}
    \caption{The prompt used for decomposing the question into atomic queries, which is used in the high-level searcher.}
    \label{fig:highlevel_decompose}
\end{figure}

\begin{figure}[h]
    \centering
    \includegraphics[width=\columnwidth]{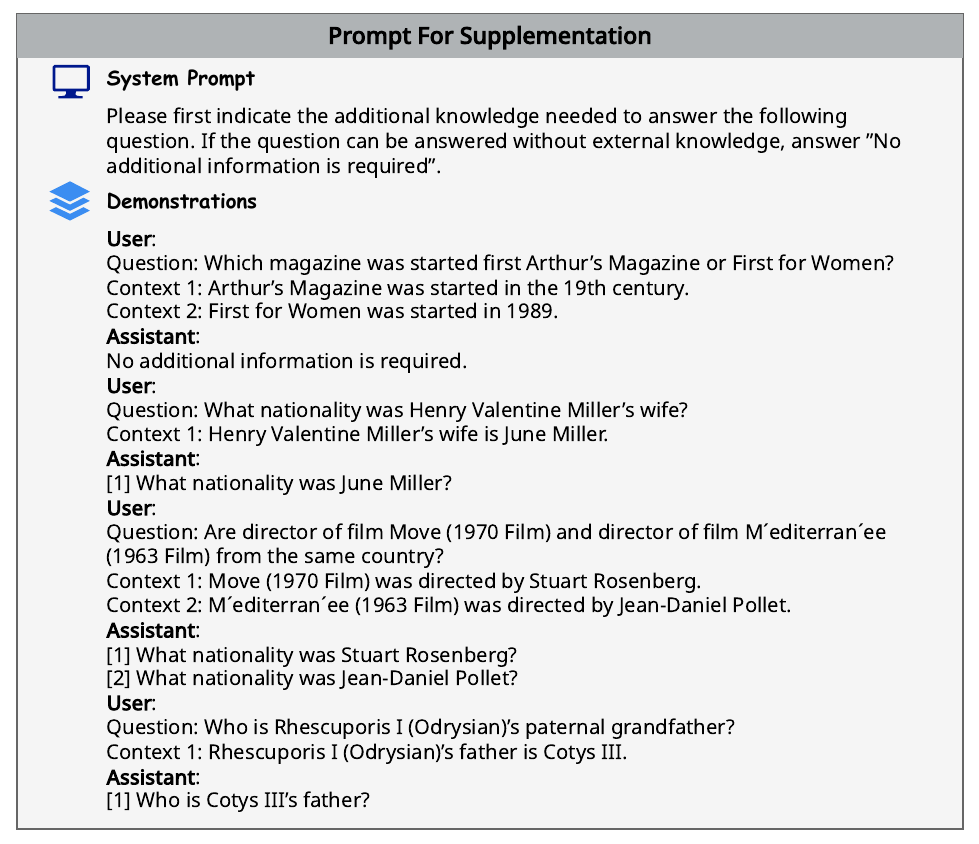}
    \caption{The prompt used to supplement the atomic queries, which is used in the high-level searcher.}
    \label{fig:highlevel_supply}
\end{figure}

\begin{figure}[t]
    \centering
    \includegraphics[width=\columnwidth]{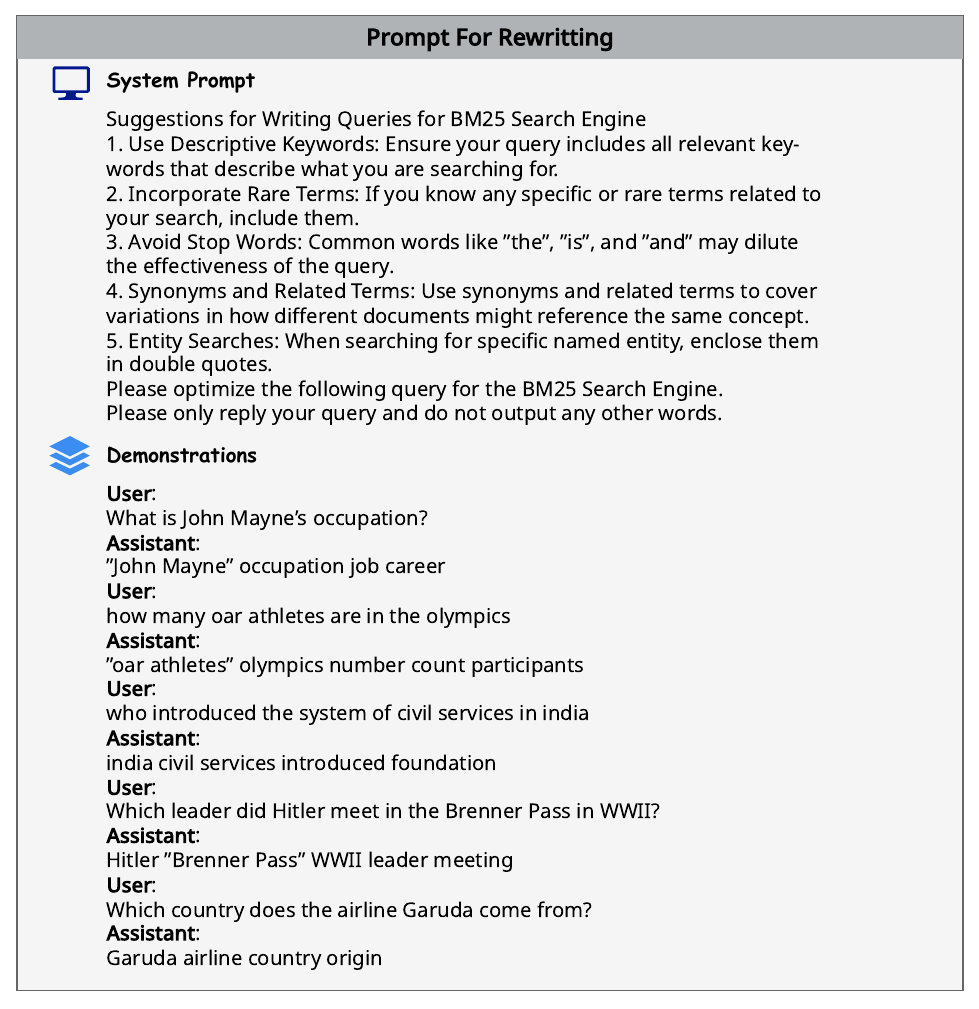}
    \caption{The prompt used to rewrite the atomic queries into keywords, which is used in the sparse searcher.}
    \label{fig:sparse_rewrite}
\end{figure}

\begin{figure}[h]
    \centering
    \includegraphics[width=\columnwidth]{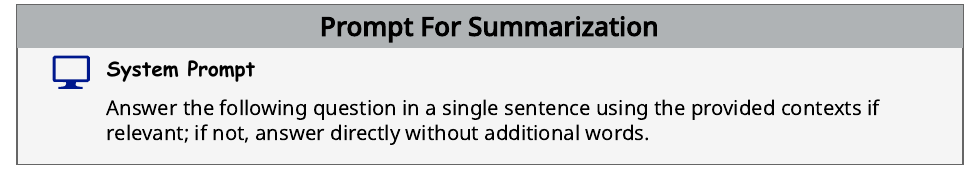}
    \caption{The prompt used to summarize the retrieved contexts, which is used in the high-level searcher.}
    \label{fig:highlevel_summary}
\end{figure}

\begin{figure}[h]
    \centering
    \includegraphics[width=\columnwidth]{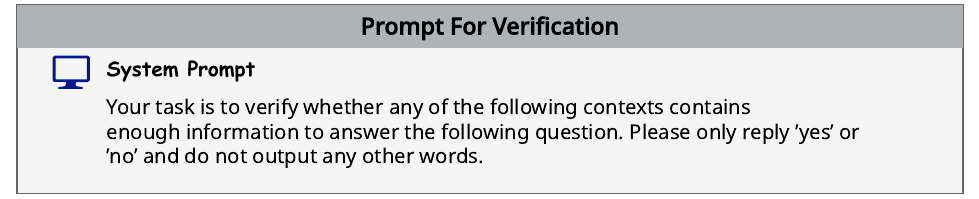}
    \caption{The prompt used to verify whether the retrieved contexts contain enough information to answer the given question, which is used in the high-level searcher, sparse searcher, and the dense searcher.}
    \label{fig:highlevel_verify}
\end{figure}

\begin{figure}[h]
    \centering
    \includegraphics[width=\columnwidth]{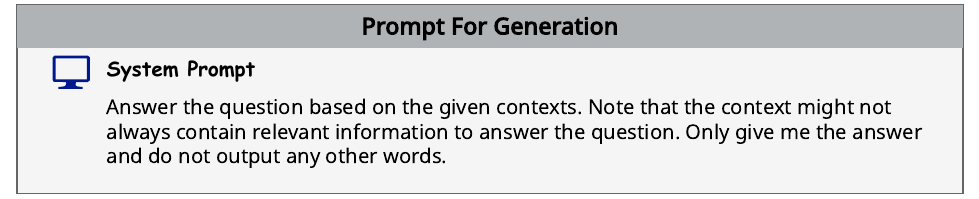}
    \caption{The prompt used for generating the final response, which is used in the generator.}
    \label{fig:generate}
\end{figure}

In all our experiments, we use the following prompts to generate responses for LevelRAG. To make the output format of the LLM fit our requirements, we use few-shot learning in some actions. The prompts are shown in Figure \ref{fig:highlevel_decompose}, \ref{fig:highlevel_supply}, \ref{fig:sparse_rewrite}, \ref{fig:highlevel_summary}, \ref{fig:highlevel_verify}, and \ref{fig:generate}.

\section{Comparison with Agent Based Methods}
To further demonstrate the effectiveness of our proposed LevelRAG, we compare it with several agent-based methods, including ReACT \cite{react}, FireACT \cite{fireact}. Following these work, we conduct experiment on the random selected 200 samples from the HotpotQA dataset. The experimental results are shown in Table \ref{tab:agent_based}. The results show that our proposed LevelRAG achieves better performance than the agent-based methods, which indicates that our proposed hierarchical searcher is more effective than the agent-based methods.

\begin{table}[h]
    \centering
    \begin{tabular}{lccc}
    \toprule
    Method               & EM              & F1              & Acc            \\
    \cmidrule(lr){1-1}  \cmidrule(lr){2-4}
    ReACT(GPT3.5-turbo)  &  25.50          &  35.82          & 32.00          \\
    ReACT(Qwen2-7B)      &  30.50          &  42.09          & 35.50          \\
    FireACT              &  24.00          &  35.35          & 29.00          \\
    LevelRAG             &  \textbf{38.00} &  \textbf{52.20} & \textbf{42.50} \\
    \bottomrule
    \end{tabular}%
    \caption{The comparison between LevelRAG and agent-based methods on 200 random selected samples from the HotpotQA dataset. \textbf{Bold} number indicates the best performance among all methods.}
    \label{tab:agent_based}
\end{table}

\section{The Completeness of the LevelRAG}
In our paper, "completeness" specifically refers to the ability of retrieving relevant contexts for any given question. We argue that using a single retriever may be insufficient due to its inherent limitations, which hybrid retrieval helps to mitigate. To demonstrate our point, we conducted an experiment on HotpotQA. The experimental results in Table \ref{tab:completeness} shows that employing hybrid retrieval can improve the completeness of the retrieval system. In addition, the performance of LevelRAG is also better than the hybrid retrieval system, which indicates that our proposed hierarchical searcher is even more effective than the hybrid retrieval system.

\begin{table}[h]
    \centering
    \begin{tabular}{lccc}
    \toprule
    Method               & EM           & F1           & Acc          \\
    \cmidrule(lr){1-1}  \cmidrule(lr){2-4}
    Sparse Searcher      &  33.21       &  44.38       & 36.07        \\
    Web Searcher         &  27.52       &  37.54       & 29.70        \\
    Dense Searcher       &  32.92       &  43.86       & 35.69        \\
    Hybrid Searcher      &  37.99       &  50.39       & 41.12        \\
    LevelRAG             &  39.32       &  39.32       & 43.78        \\
    \bottomrule
    \end{tabular}%
    \caption{The completeness of LevelRAG and other methods on HotpotQA dataset.}
    \label{tab:completeness}
\end{table}

\end{document}